\documentclass[journal]{IEEEtran}\usepackage{amsmath,amsfonts,amssymb}
\usepackage{amsthm}
\usepackage{adjustbox}
\usepackage{mathrsfs}
\usepackage{algorithmicx}
\usepackage{algpseudocode}
\usepackage{algorithm}
\usepackage{array}
\usepackage[caption=false,font=normalsize,labelfont=sf,textfont=sf]{subfig}
\usepackage{textcomp}
\usepackage{stfloats}
\usepackage{url}
\usepackage{verbatim}
\usepackage{graphicx}
\usepackage{cite}
\usepackage{svg}
\usepackage{multirow}
\usepackage{booktabs}
\usepackage{xcolor}
\usepackage{listings}
\usepackage[T1]{fontenc}

\newcommand{\citep}[1]{\cite{#1}}
\newcommand{\citet}[1]{\cite{#1}}
\theoremstyle{plain}

\theoremstyle{definition}

\begin{document}

\title{Efficient On-Board Processing of Oblique UAV Video for Rapid Flood Extent Mapping}
\author{Vishisht~Sharma,
        Sam~Leroux,
        Lisa~Landuyt,
        Nick~Witvrouwen,
        Pieter~Simoens% 
\thanks{This work was supported by the Belgian Science Policy Office (BELSPO) via the Stereo IV Program, grant number SR/00/415 (FLOWS). (Corresponding author: Vishisht Sharma.)}%
\thanks{V. Sharma is with IDLab, Department of Information Technology, Ghent University - imec, 9052 Gent, Belgium, and also with the Environmental Intelligence Unit, VITO (Flemish Institute for Technological Research), 2400 Mol, Belgium (e-mail: vishisht.sharma@ugent.be).}%
\thanks{S. Leroux and P. Simoens are with IDLab, Department of Information Technology, Ghent University - imec, 9052 Gent, Belgium (e-mail: sam.leroux@ugent.be; pieter.simoens@ugent.be).}%
\thanks{L. Landuyt and N. Witvrouwen are with the Environmental Intelligence Unit, VITO (Flemish Institute for Technological Research), 2400 Mol, Belgium (e-mail: lisa.landuyt@vito.be; nick.witvrouwen@vito.be).}%
\thanks{Manuscript received December 31, 2025.}}
% {Anonymous Authors: Efficient On-Board Processing of Oblique Aerial Video}

\maketitle

\begin{abstract}
Effective disaster response relies on rapid disaster response, where oblique aerial video is the primary modality for initial scouting due to its ability to maximize spatial coverage and situational awareness in limited flight time. However, the on-board processing of high-resolution oblique streams is severely bottlenecked by the strict Size, Weight, and Power (SWaP) constraints of Unmanned Aerial Vehicles (UAVs). The computational density required to process these wide-field-of-view streams precludes low-latency inference on standard edge hardware. To address this, we propose Temporal Token Reuse (TTR), an adaptive inference framework capable of accelerating video segmentation on embedded devices. TTR exploits the intrinsic spatiotemporal redundancy of aerial video by formulating image patches as tokens; it utilizes a lightweight similarity metric to dynamically identify static regions and propagate their precomputed deep features, thereby bypassing redundant backbone computations. We validate the framework on standard benchmarks and a newly curated Oblique Floodwater Dataset designed for hydrological monitoring. Experimental results on edge-grade hardware demonstrate that TTR achieves a 30\% reduction in inference latency with negligible degradation in segmentation accuracy (<<0.5\% mIoU). These findings confirm that TTR effectively shifts the operational Pareto frontier, enabling high-fidelity, real-time oblique video understanding for time-critical remote sensing missions.
\end{abstract}


\section{Introduction}\label{sec:intro}
\IEEEPARstart{T}{he} widespread adoption of Unmanned Aerial Vehicles (UAVs) in applications ranging from disaster response to precision agriculture \citep{Maggiori2017Inria} has created an unprecedented demand for real-time, on-board data analysis. However, processing UAV-captured video presents a unique and often overlooked computational dilemma, born from a fundamental conflict between the data's characteristics and the operational nature of modern deep learning models. On one hand, the video is frequently captured at high resolutions, e.g., 1920x1080 (Full HD) or even 4K, to ensure that fine-grained details are preserved for accurate analysis. This high pixel count inherently demands significant processing power and memory bandwidth for every single frame. On the other hand, this data is also intrinsically characterized by vast temporal redundancy \citep{Wu2017Spatial}. In a typical mission, a UAV may execute prolonged flight patterns, including hovering for surveillance, carrying out systematic pans for mapping, or orbiting specific points of interest, such as during disaster response scenarios. These maneuvers cause large portions of the observed scene large agricultural fields, static water bodies, unchanging forests, or stationary buildings to remain perceptually identical or nearly identical across tens, if not hundreds, of consecutive frames.

This redundancy is impacted by the brute-force nature of modern deep neural networks. While powerful, these architectures, by design, treat each incoming frame as an independent problem. They dictate that every pixel, or every patch in the case of a Transformer-based network \citep{xie2021segformer}, must be processed through the entire computational graph, from the initial layers to the final prediction head. This exhaustive processing is ill-suited for video streams where the temporal correlation between frames is high, often for multiple consecutive frames. 

% FIGURE 1: Introduction Example
\begin{figure}[!t]
\centering
\includegraphics[width=\columnwidth]{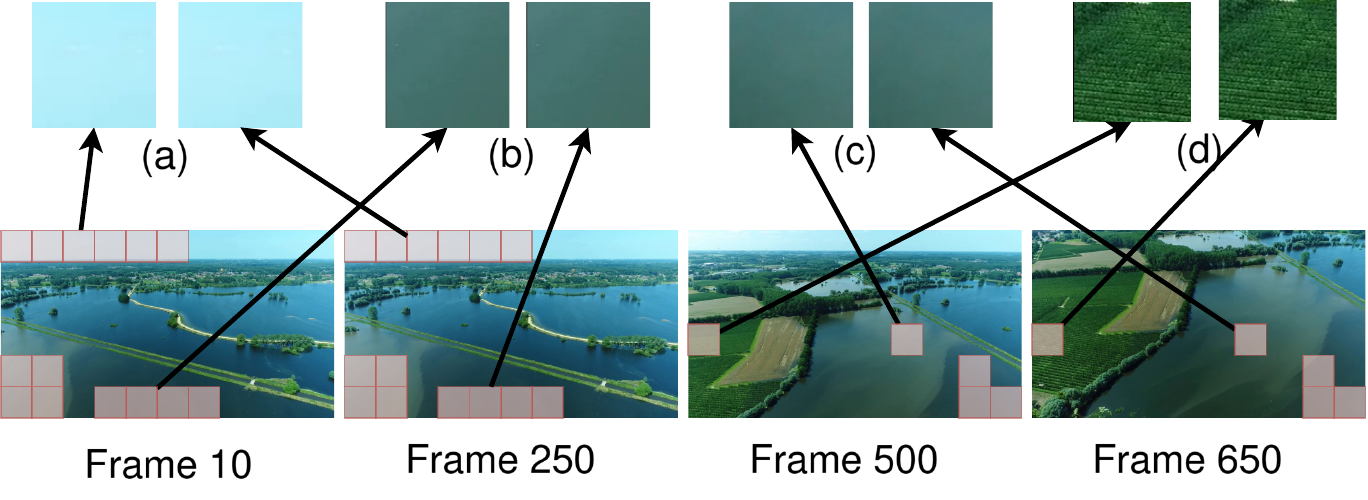}
\caption{We visualize four pairs of patches to demonstrate high self-similarity over time. For instance, the sky patch in (a) and the water patch in (b) remain almost identical between Frame 10 and Frame 250. Similarly, patches for water (c) and vegetation (d) show negligible change over hundreds of frames despite camera motion. Our TTR framework leverages a high cosine similarity score between such patches to bypass their re-computation.}
\label{fig:similar_patches}
\end{figure}

Figure \ref{fig:similar_patches}, for instance, demonstrates this correlation by showing how patches of water and vegetation maintain high self-similarity even with significant camera motion and across a gap of over 200 frames. Consequently, processing these large, unchanging static regions with a computationally expensive model constitutes an inefficient use of computational cycles. This inefficiency is not a minor technical issue; it results directly into operational limitations: shorter flight times due to the high power draw of the on-board processing \citep{Sze2017Efficient, Canziani2017Analysis}, a data surge that slows down post-mission analysis on the ground, and a bottleneck for applications that demand immediate situational awareness, such as search and rescue or flood mapping, where latency can render data obsolete the moment it is processed.

This disparity, where some parts of a scene are highly dynamic (e.g., moving vehicles, flowing water, or the leading edge of an ongoing disaster) while the vast majority remains static, calls for a paradigm shift towards ``Adaptive Computation'' \citep{Figurnov2017Spatially, Wu2018BlockDrop}. This principle rejects the one-size-fits-all, uniform processing model and instead suggests a more intelligent, dynamic allocation of computational effort, precisely tailored to the spatiotemporal complexity and novelty of the input data. An adaptive system built on this paradigm operates with a nuanced understanding of the scene's dynamics. For instance, it would allocate the maximum computational budget to track and segment high-priority, moving objects or rapidly changing land cover. These are the regions where accuracy is the most important and where new information resides. Conversely, for the large, static, and unaffected areas of the scene the sky or a distant mountain range it would expend only the minimal computational effort required to verify their static nature, before reusing previously computed results.

This processing strategy is not just an optimization; it is a requirement for enabling the potential of UAV applications in real-world scenarios. By focusing finite computational resources where they are needed most, this approach makes it possible to achieve two critical goals simultaneously: maintaining high fidelity and accuracy in the dynamic regions of interest, while improving overall system efficiency. Our method bridges the gap between computationally demanding, high-accuracy models and the latency requirements of power-constrained edge devices, enabling their deployment in real-world missions.

The primary contribution of this paper is the proposed Temporal Token Reuse (TTR) framework, which directly targets the temporal redundancy. It introduces a computationally inexpensive mechanism to cache and reuse the deep feature representations of static scene regions. By identifying and reusing the pre-computed features of tokens that have not visually changed since the previous frame, TTR reduces the computational load on the main segmentation backbone. This ensures that processing power is precisely focused only where it is needed most: on the dynamic and actively changing parts of the scene. This adaptive approach is crucial for optimizing the performance-to-power consumption ratio, enabling the potential of deploying high-accuracy, high-fidelity segmentation models on power- and compute-constrained aerial platforms.

Given our approach of treating image patches as tokens, a Transformer-based architecture might seem like the obvious choice. However, we deliberately diverge from this path due to the practical realities of on-board UAV processing. The very strength of Transformers, their ability to model global relationships via self-attention, is also their primary drawback in this context, as its quadratic complexity is exceptionally demanding for high-resolution, real-time video on power-limited hardware. In contrast, Convolutional Neural Networks (CNNs) offer a more pragmatic and tenable foundation, with a rich history of lightweight designs optimized for edge deployment. Additionally, emerging CNN frameworks grant us the Transformer-like granular control over distinct image regions, a necessity for our temporal reuse strategy while retaining the inherent efficiency of convolutional neural networks. By building upon a CNN backbone, our approach represents a deliberate trade-off: we prioritize the practical feasibility of deploying a high-fidelity, adaptive system in a real-world operational environment over adherence to the latest, albeit computationally expensive, architectural trends.

%-----------------------------------------------------------------------------------------------------------------------------------------------------------------------

\section{Related Work}\label{sec:related_research}

Our work builds upon two domains of research: video segmentation and efficient deep learning. The challenge of deploying deep neural networks on resource-constrained platforms, such as Unmanned Aerial Vehicles (UAVs), has catalyzed significant innovation in model optimization. As surveyed by \citet{Sze2017Efficient}, these efforts can be broadly divided into static model compression, efficient network design, and dynamic inference strategies.

\subsection{Efficient Network Architectures}
Static methods aim to create a single, efficient model before deployment. This includes techniques like network pruning, which removes redundant weights or channels \citep{LeCun1990Optimal}, a field that continues to evolve with methods for structured pruning that can be reconfigured without retraining \citep{Liu2023Unstructured} and data-free pruning techniques for large-scale models \citep{Zhang2023DataFree}; quantization, which reduces the numerical precision of network parameters \citep{Wu2016Quantized}, with modern post-training methods now capable of accurately quantizing large vision models with minimal data \citep{Li2023OmniQuant}; and knowledge distillation, where a smaller student network is trained to mimic the output of a larger, more powerful teacher network \citep{Hinton2015Distilling}, with recent research exploring how to distill knowledge from multiple diverse teachers effectively \citep{Beyer2022Knowledge}. Furthermore, a significant body of work has focused on designing inherently efficient CNN architectures from the ground up, such as MobileNetV2 \citep{Sandler2018MobileNetV2}, ShuffleNetV2 \citep{Ma2018ShuffleNetV2}, and ERFNet \citep{Romera2018ERFNet}, which utilize optimized operations like depth-wise separable convolutions and channel shuffling. This trend continues with modern designs that employ large kernel convolutions for efficiency at scale \citep{Ding2022Scaling} and lightweight Vision Transformers tailored for mobile applications \citep{Mehta2022MobileViTv2}. While these approaches successfully reduce the overall computational footprint (as measured in FLOPs) and memory requirements, they share a fundamental limitation: they still apply the same uniform computational graph to every pixel of every input frame. This makes them incapable of exploiting the massive spatio-temporal redundancy inherent in video data.

\subsection{Frame-Level and Motion-Based Temporal Methods}
To address temporal redundancy at a coarse frame level, a straightforward strategy is the Keyframe-based approach. These techniques perform full, computationally expensive segmentation on sparsely selected keyframes and then use a faster method for the intermediate delta frames. Early methods relied on simple interpolation, but more advanced techniques have been developed to propagate and aggregate deep features from keyframes to subsequent frames, improving accuracy \citep{Zhu2017DFF}. While simple in principle, this global decision-making process presents a stark trade-off between efficiency and temporal fidelity. A fixed, low keyframe rate drastically increases the risk of missing transient events, which has led to research on adaptive frame selection strategies that dynamically choose keyframes based on scene content \citep{Wu2018AdaFrame}. More recent works have framed this as a policy-learning problem, where a lightweight network learns to dynamically skip frames to maximize both efficiency and accuracy for a given computational budget \citep{Gundavarapu2023Lets}. Further extending this, recent learning-based paradigms like co-training have been proposed to leverage the information from unlabeled intermediate frames better, improving overall temporal consistency \citep{Paul2021CoTraining}.

A more sophisticated and dominant paradigm is Feature Propagation with Optical Flow. This method calculates a dense optical flow field to warp feature maps from a keyframe to the current frame, thus maintaining temporal consistency and improving segmentation accuracy over time \citep{Li2018VSS}. However, dense optical flow estimation is itself a computationally intensive task, which can offset the intended efficiency gains \citep{ALFARANO2024104160}. More critically, it is prone to errors in challenging but common scenarios such as occlusions, textureless surfaces, and illumination changes, which are prevalent in aerial footage and can degrade segmentation quality \citep{Gao_2024}. Recognizing these limitations, recent research has focused on mitigating flow-induced errors, for example, by using attention mechanisms to blend features selectively and leveraging uncertainty estimation to identify and refine unreliable warped regions \citep{Wang2022AFBURR}. State-of-the-art approaches now seek to replace explicit flow entirely with more robust, learnable correspondence matching at the feature level, which can better handle large displacements and appearance changes \citep{Zhou2023Rethinking}. Despite these advances, both keyframing and motion warping methods apply their temporal logic uniformly across the entire frame, lacking the fine-grained adaptability to handle scenes with spatially varying dynamics.

\subsection{Spatially-Dynamic Computation}
Parallel to temporal efficiency methods, a highly relevant line of research has focused on exploiting Spatial Sparsity within a single frame. The guiding principle, as articulated by \citet{Li2017Difficulty}, is that not all pixels are created equal. This has led to a variety of dynamic networks that allocate computation unevenly across the spatial domain \citep{Wu2017Spatial}. One major approach involves the selective processing of image regions. For instance, SBNet \citep{Ren2018SBNet}, which inspired the SegBlocks framework, uses a lightweight policy network to identify a sparse subset of image blocks for processing. This concept has been modernized and extensively explored in Vision Transformers, which employ strategies like adaptive token selection to process a variable number of tokens based on input complexity \citep{Liang2023EViT}, token merging to consolidate redundant features \citep{Bolya2023Token}, and token routing to focus computation on the most salient image regions \citep{Tang2023Dynamic}.

Other methods adapt the network's structure or depth dynamically. SkipNet \citep{Wang2018SkipNet} and BlockDrop \citep{Wu2018BlockDrop} learn to conditionally bypass entire residual blocks for simpler image regions, creating adaptive inference graphs tailored to the input \citep{Veit2018Adaptive}. This principle is now central to dynamic Vision Transformers that can conditionally skip entire layers or self-attention blocks based on intermediate feature representations \citep{Cong2023PoWER-BERT}. A more granular approach, dynamic convolutions \citep{Verelst2020Dynamic}, which was further refined to be more hardware-friendly \citep{Li2022Dynamic}, exploits spatial sparsity directly within the convolutional operation itself. Another variant, proposed by \citet{Figurnov2017Spatially}, adapts computation time per pixel, halting computation early for regions classified with high confidence, a concept now successfully applied to Vision Transformers through layer-wise early exiting mechanisms \citep{Sun2023Implicit}. These methods have proven highly effective for efficient high-resolution image processing \citep{Wu2020Patch, Zhao2018ICNet}. However, their core limitation is that their focus is strictly intra-frame; they re-evaluate the spatial complexity of the entire scene from scratch for every single video frame, thereby failing to capitalize on the most significant source of redundancy in aerial video surveys: temporal staticity.

In stark contrast, our Temporal Token Reuse (TTR) framework is conceived to address these shortcomings directly. It synthesizes the fine-grained, patch-level adaptability of spatial sparsity methods like SBNet \citep{Ren2018SBNet} and extends this principle into the temporal dimension. Instead of the costly and error-prone estimation of optical flow, TTR employs a computationally cheap similarity metric to identify static regions over time precisely. Unlike coarse keyframe-based methods, TTR operates at a local level, allowing it to preserve high temporal fidelity for dynamic objects while aggressively saving computation in static areas. By caching and reusing features for unchanged regions, TTR elegantly avoids redundant processing, ensuring that finite computational resources are focused exclusively where they are needed most. This makes it a uniquely flexible, robust, and efficient solution for the demands of real-time segmentation on video sequences.

\section{Methodology}\label{sec:methodology}

\subsection{SegBlocks: A Foundation for Adaptive CNN Processing}\label{subsec:segblocks}
The SegBlocks \citep{DBLP:journals/corr/abs-2011-12025} framework represents a critical step towards making CNNs more adaptive. It partitions an input image into a grid of blocks and dynamically adjusts the processing resolution based on spatial complexity. This innovation allows a CNN to focus its computational resources by processing detailed or complex regions at a high resolution while using a lower, more computationally efficient resolution for simpler areas. We leverage this patch-based architecture as the fundamental building block for our temporal optimization, as it provides the necessary mechanism to process image regions independently.

A core challenge in any sparse, patch-based CNN framework is preserving spatial context and continuity. The effectiveness of CNNs relies on their receptive field, which allows neurons to gather information from a local neighborhood. In a standard dense convolution, this neighborhood is always available. However, in a sparse framework where only a subset of active patches are processed, the filters at the boundaries of these patches face a problem: their receptive fields extend into neighboring inactive patches whose features for the current network stage have not been computed. This lack of contextual information breaks the feature sharing across the frame, leading to artifacts visible as discontinuities or incorrect classifications along the seams of the processed blocks, thus degrading the final segmentation quality.

To resolve this dilemma, SegBlocks employs a specialized operation known as Boundary Feature Copying, implemented within a BlockCopy \citep{DBLP:journals/corr/abs-2011-12025} module. The primary function of this module is to reconstruct the necessary local neighborhood for each active patch before it undergoes convolution. After a sparsity mask indicates which patches are active, the BlockCopy module prepares the input for the convolutional layer by creating a temporary, larger patch padded with features from its immediate one-pixel-deep neighborhood, i.e., its 8 neighbors. The key innovation lies in how this padding is sourced:

\begin{itemize}
\item If a neighboring patch is processed at the same resolution (i.e., it is also 'active'), its features are copied directly from the current stage's input feature map.
\item If a neighboring patch is processed at a lower resolution, its smaller feature map is retrieved and then upsampled, e.g., using nearest-neighbor interpolation to match the resolution of the active patch before being copied into the padding region.
\end{itemize}

Once this contextually aware padded patch is constructed, the respective convolutional block is applied to it. The receptive field of the convolutions can now operate correctly, even at the edges. This mechanism seamlessly stitches the sparsely computed results together, preserving the integrity of the convolutional operation while enabling the computational savings of adaptive-resolution processing.

\subsection{Temporal Token Reuse (TTR)}\label{subsec:ttr_mechanism}

While SegBlocks optimizes computation within a single frame by varying spatial resolution, our primary contribution, Temporal Token Reuse (TTR), extends this concept into the temporal domain for video processing. We modify the core mechanism: instead of merely reducing the processing resolution, TTR skips computation entirely for patches that have not changed significantly from the previous frame. By identifying these temporally redundant tokens, we reuse the rich, deep feature maps computed for the previous frame, achieving substantial efficiency gains in scenes with static elements.

% FIGURE 2: Methodology Overview
\begin{figure*}[!t]
\centering
% --- Top Subfigure (a) ---
\subfloat[A high-level conceptual overview of the proposed framework. The input is processed collaboratively by the Temporal Token Reuse (TTR) Module and the main CNN + SegBlocks Backbone, which generates the final segmentation mask.]{\includegraphics[width=\textwidth]{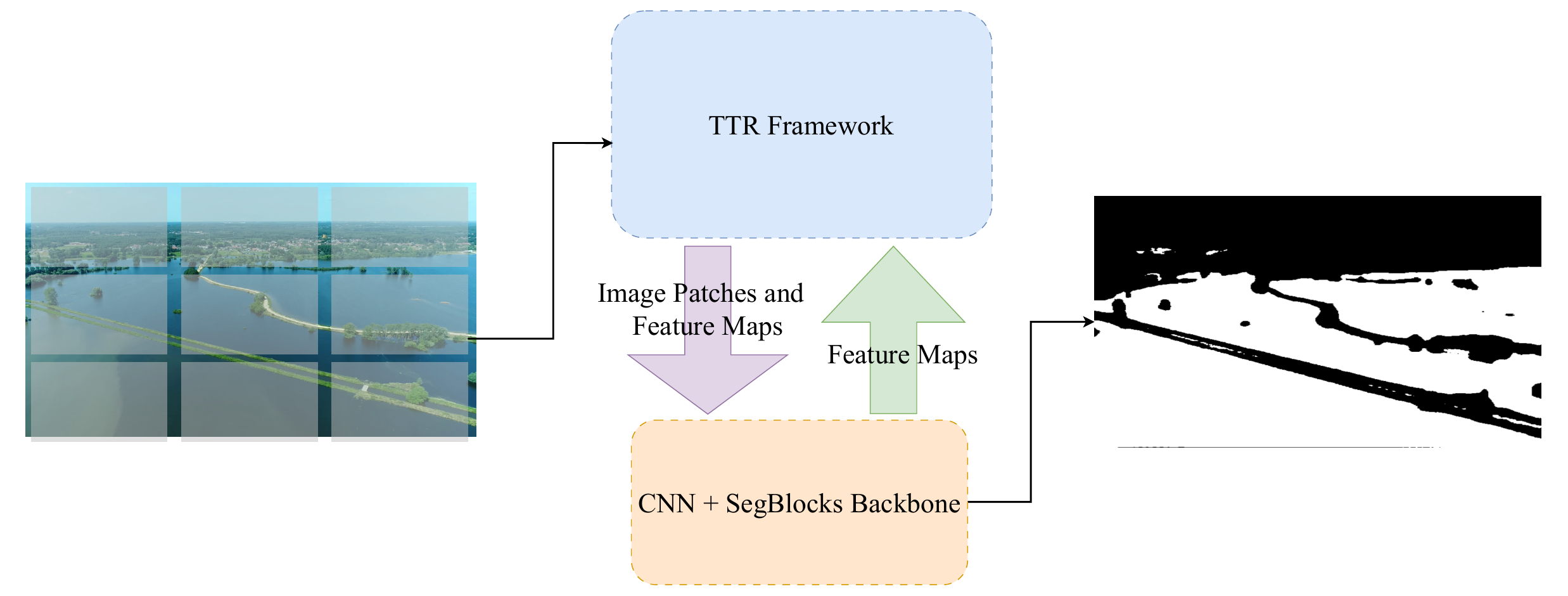}%
\label{fig:TTR_high_level}}
\hfil
\par
% --- Left Subfigure (b) ---
\subfloat[Temporal Token Reuse Framework]{\includegraphics[width=0.49\textwidth]{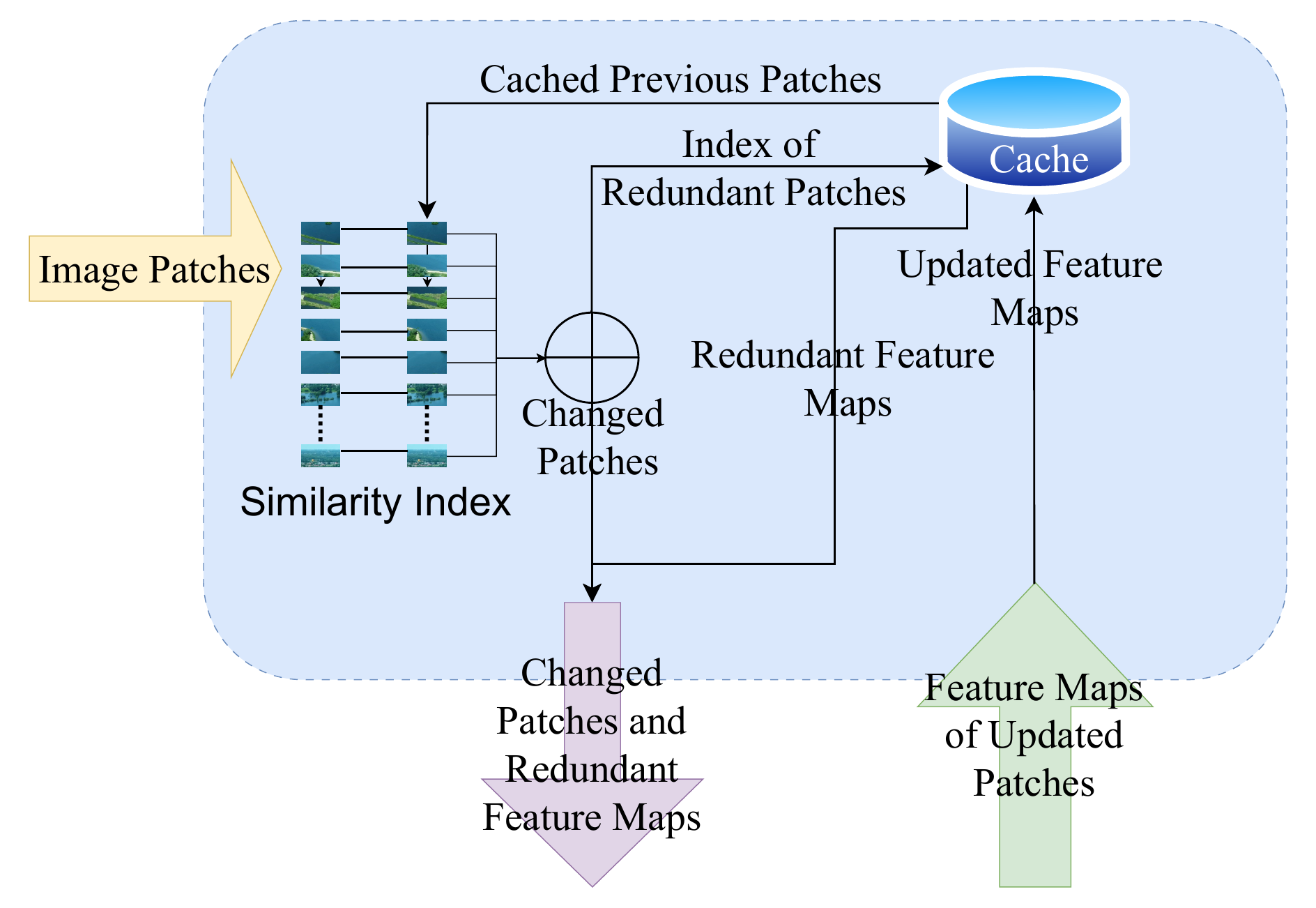}%
\label{fig:TTR_detailed_flow}}
\hfil
% --- Right Subfigure (c) ---
\subfloat[CNN + Segblocks]{\includegraphics[width=0.49\textwidth]{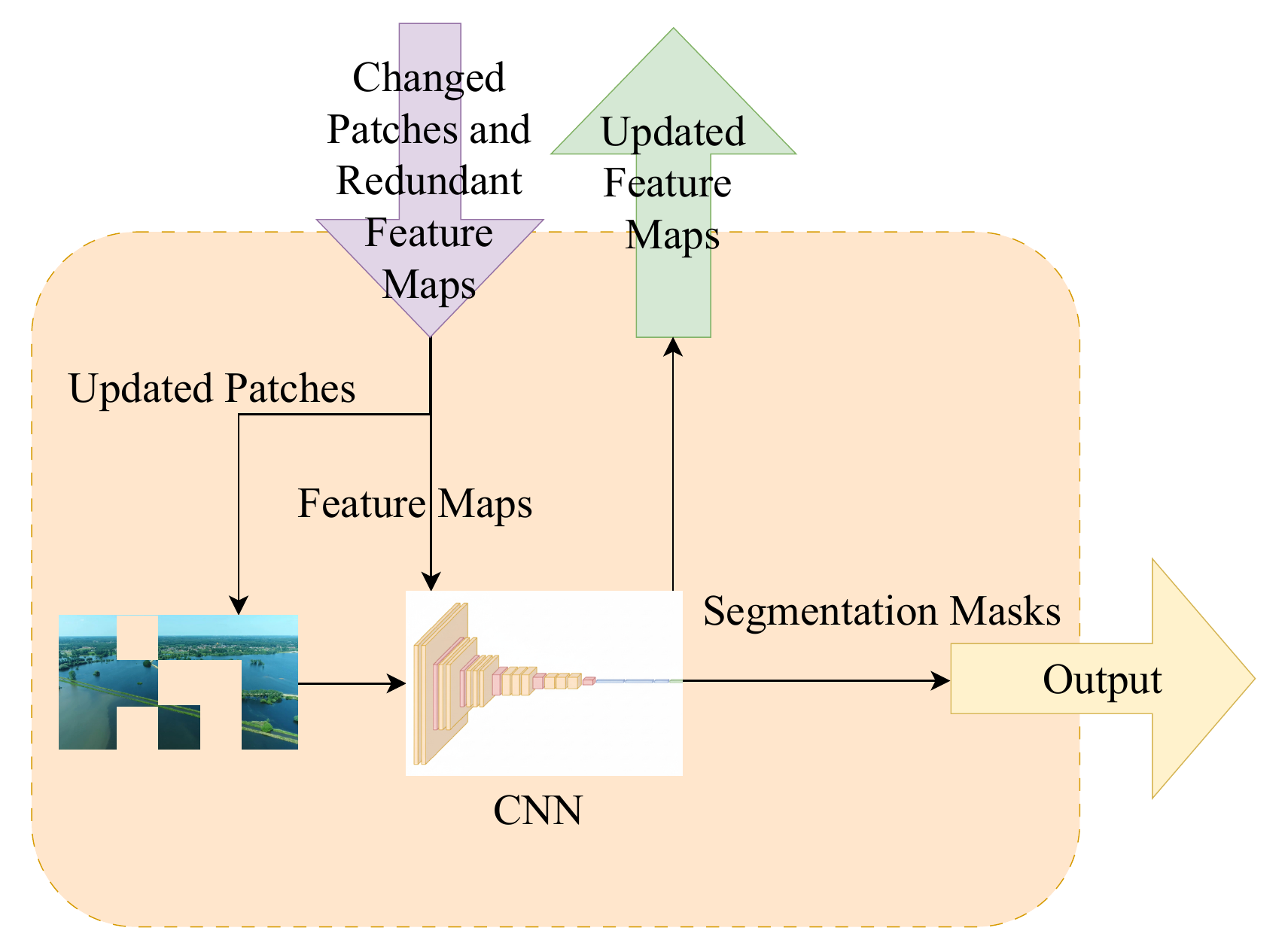}%
\label{fig:SEG_detailed_flow}}
\caption{A detailed operational diagram of the TTR framework. The data flow bifurcates based on patch similarity: (1) Changed patches undergo full CNN processing, with the new features used to update the cache. (2) For redundant patches, features are retrieved directly from the cache, bypassing the CNN. The final mask is assembled from both new and reused features.}
\label{fig:framework_comparison}
\end{figure*}

The architecture and operational flow of this process are detailed in Figure \ref{fig:framework_comparison}. The TTR process for each frame is formalized into two key stages, detailed in Algorithms \ref{alg:ttr_mask_generation} and \ref{alg:ttr_forward_pass}.

% --- ALGORITHM 1: Fixed Gap and Width ---
\begin{algorithm}[!t] % changed [H] to [!t] to fix the white space gap
\small % Reduces font size to fit column
\caption{TTR Sparsity Mask Generation}
\label{alg:ttr_mask_generation}
\begin{algorithmic}[1]
\Require Current Frame $I_t$
\Require Previous Frame's Patch Cache: 
    \Statex \hskip\algorithmicindent \texttt{PrevPatchCache} % Manually broken line
\Require Cosine similarity threshold $\tau$
\Ensure Sparsity Mask: \texttt{SparsityMask}
\Ensure Updated Cache: \texttt{CurrPatchCache}

\State Let $P_t$ be the set of patches $\{p_{t,i}\}$ from frame $I_t$
\State Initialize empty \texttt{SparsityMask}
\For{each patch $p_{t,i}$ in $P_t$}
    \State $p_{t-1,i} \leftarrow \text{PrevPatchCache}[i]$
    % Break long equation line if needed
    \If{$\text{CosSim}(p_{t,i}, p_{t-1,i}) \leq \tau$} 
        \State $\text{SparsityMask}[i] \leftarrow \text{ACTIVE}$
    \Else
        \State $\text{SparsityMask}[i] \leftarrow \text{REDUNDANT}$
    \EndIf
    \State $\text{CurrPatchCache}[i] \leftarrow p_{t,i}$
\EndFor
\State \Return \texttt{SparsityMask}, \texttt{CurrPatchCache}
\end{algorithmic}
\end{algorithm}

First, a lightweight decision mechanism determines which image regions require re-computation. For each incoming frame $I_t$, we compare its patches against those from the preceding frame $I_{t-1}$ using the Cosine Similarity Index. Each image patch (containing raw RGB values) is flattened into a vector, and its similarity to the corresponding patch in the previous frame is calculated as:
\begin{equation}
\text{Similarity}(p_i) = \frac{p_i(t) \cdot p_i(t-1)}{|p_i(t)| |p_i(t-1)|} \label{eq1}
\end{equation}
We define a high similarity threshold, $\tau$. If $\text{Similarity}(p_i) > \tau$, the block is classified as static (REDUNDANT). Otherwise, it is considered dynamic (ACTIVE). We chose cosine similarity because it is inherently robust to uniform brightness changes. This logic for generating a SparsityMask based on inter-frame similarity is formally presented in Algorithm \ref{alg:ttr_mask_generation}.
% --- ALGORITHM 2: Fixed Gap and Width ---
\begin{algorithm}[!t] % changed [H] to [!t]
\small % Reduces font size
\caption{TTR-Enhanced Forward Pass}
\label{alg:ttr_forward_pass}
\begin{algorithmic}[1] % Changed [2] to [1] for standard numbering
\Require Frame $I_t$, CNN \texttt{Model} with BlockSkip
\Require \texttt{SparsityMask} from Algorithm \ref{alg:ttr_mask_generation}
\Require Prev. Feature Caches: \texttt{PrevFeatCaches}
\Ensure Mask $M_t$, Updated Caches: \texttt{CurrFeatCaches}

\State Initialize \texttt{CurrFeatCaches}
\State $x \leftarrow \text{FeatureMap}(I_t)$
\For{each stage $l$ in \texttt{Model.Stages}}
    \State $x_{\text{act}} \leftarrow \text{GetActive}(x, \text{SparsityMask})$
    
    % Manually break this long line
    \State $x_{\text{pad}} \leftarrow \text{BlockSkip}(x_{\text{act}}, \text{PrevFeatCaches}[l],$
    \Statex \hskip\algorithmicindent \hskip\algorithmicindent $\text{SparsityMask})$

    \State $f_{\text{new}} \leftarrow \text{Model.Stages}[l](x_{\text{pad}})$

    % Manually break this long line
    \State $x_{\text{reuse}} \leftarrow \text{GetRedundant}(\text{PrevFeatCaches}[l],$ 
    \Statex \hskip\algorithmicindent \hskip\algorithmicindent $\text{SparsityMask})$

    \State $x \leftarrow \text{Assemble}(f_{\text{new}}, x_{\text{reuse}}, \text{SparsityMask})$
    \State $\text{CurrFeatCaches}[l] \leftarrow x$
\EndFor

\State $M_t \leftarrow \text{Model.FinalLayer}(x)$
\State \Return $M_t$, \texttt{CurrFeatCaches}
\end{algorithmic}
\end{algorithm}

Once the SparsityMask is generated, it governs the data flow within the model's forward pass. A critical component enabling this sparse processing is our BlockSkip module, a modification of the original BlockCopy mechanism tailored for temporal reuse. Its purpose is to preserve spatial context across the sparsely computed feature maps. For each ACTIVE patch about to be processed by a convolutional layer, BlockSkip reconstructs its local neighborhood by padding it with features from its neighbors. If a neighboring patch is also ACTIVE, its features are sourced from the current computational graph. However, if a neighbor is REDUNDANT, its features are retrieved directly from the corresponding location in the previous frame's feature cache for that specific layer. By simulating a dense feature map locally, BlockSkip allows standard convolutional layers to operate correctly on sparse inputs, stitching together newly computed and reused features and preventing blocking artifacts that would otherwise degrade segmentation quality. This layer-wise process is detailed in Algorithm \ref{alg:ttr_forward_pass}.

% FIGURE 3: Threshold Impact
\begin{figure}[!t]
\centering
\includegraphics[width=\columnwidth]{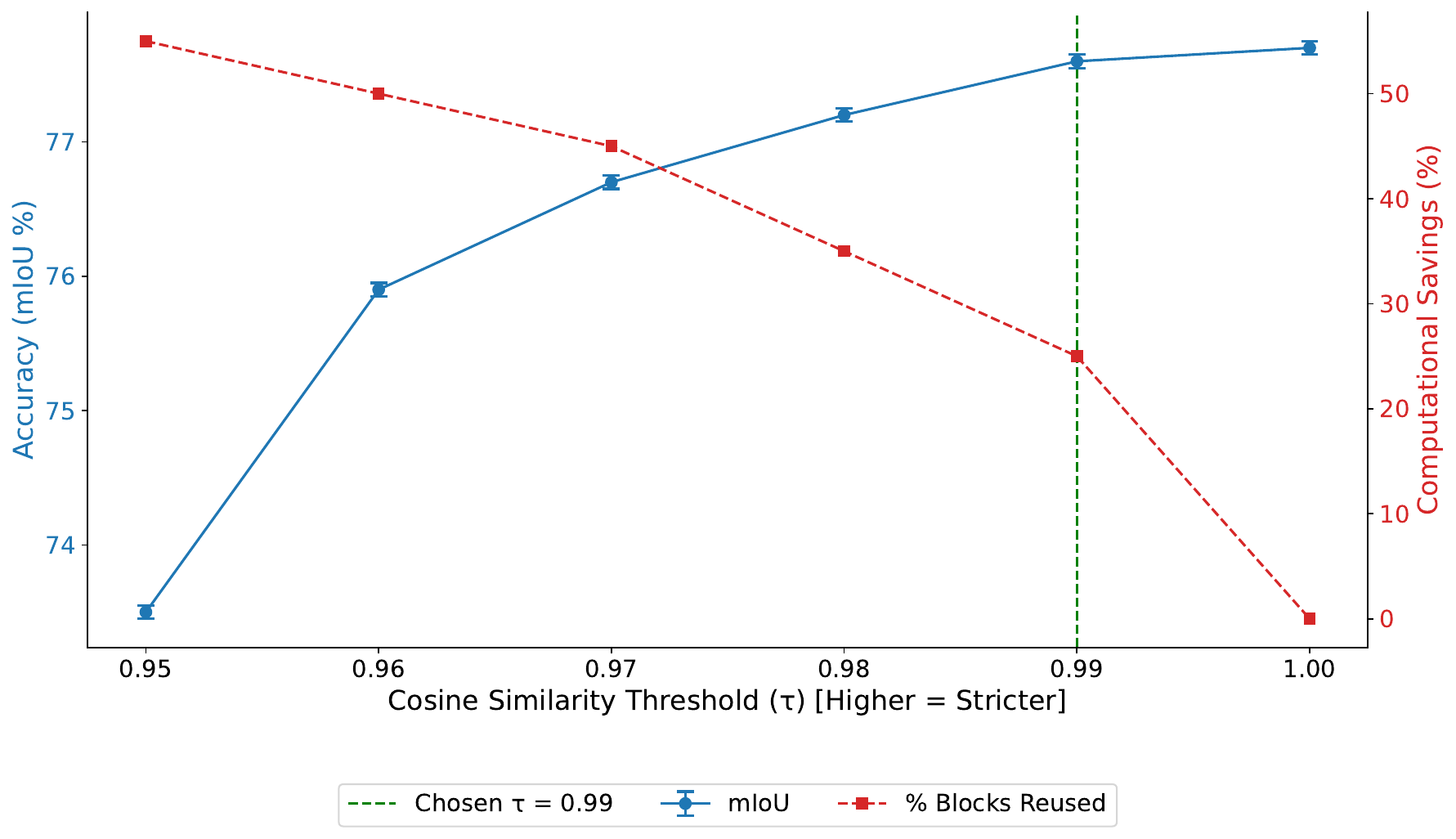}
\caption{The trade-off between segmentation accuracy (mIoU, blue) and computational savings (\% Blocks Reused, red). As the threshold becomes stricter (moving right), accuracy improves at the cost of reduced savings. We selected $\tau=0.99$ as our operating point, offering a strong balance between high performance and significant computational reduction.}
\label{fig:threshold_impact}
\end{figure}

The feature cache aids the computational savings of TTR. This is not a single cache of final predictions, but a collection of deep feature tensors, one for each stage of the network. This process forms a continuous cycle: the fully assembled feature maps (CurrFeatureCaches) generated during the forward pass of frame $I_t$ become the PrevFeatureCaches for the subsequent frame $I_{t+1}$, ensuring the system is always ready for the next comparison. This strategic coupling of a feature cache with sparse input processing results in an efficient and adaptive computational workflow. This design provides a mechanism for runtime control, allowing the computational budget to be modulated through the threshold parameter $\tau$.

The impact of the threshold hyperparameter $\tau$ is analyzed in Figure \ref{fig:threshold_impact}. It acts as a control knob, allowing a user to balance the system between peak efficiency and maximum accuracy dynamically. For a rapid initial damage assessment, an operator could use a lower $\tau$ to maximize speed. Conversely, for a detailed survey mission, a higher $\tau$ would ensure higher fidelity. This highly favorable trade-off, where a significant boost in FPS is achieved for a small drop in mIoU, makes TTR a compelling solution for resource-constrained devices.

\section{Results}\label{sec:results}

\subsection{Datasets}\label{subsec:datasets}

To align with the operational requirements of real-time Earth observation, our evaluation prioritizes domain-specific efficacy in disaster monitoring before establishing broader architectural robustness. We commence with a targeted analysis using the FloodNet dataset, a standardized benchmark for post-disaster damage assessment, alongside our novel Floodwater dataset. While FloodNet validates the model's fidelity in classifying complex disaster features against community standards, the Floodwater dataset curated for real-time aerial scouting demonstrates TTR's efficiency in the precise operational oblique video scenarios for which it was designed. Subsequently, to confirm the framework's generalizability across varying sensing conditions, we extend our evaluation to disparate public benchmarks: the aerial UAVid dataset, assessing performance in dense urban environments, and the ground-level A2D2 dataset, serving as a stress test for high-motion, low-redundancy limits.

\subsubsection{The UAVid Dataset}

% FIGURE 4: UAVid Examples
\begin{figure}[!t]
\centering
\includegraphics[width=\columnwidth]{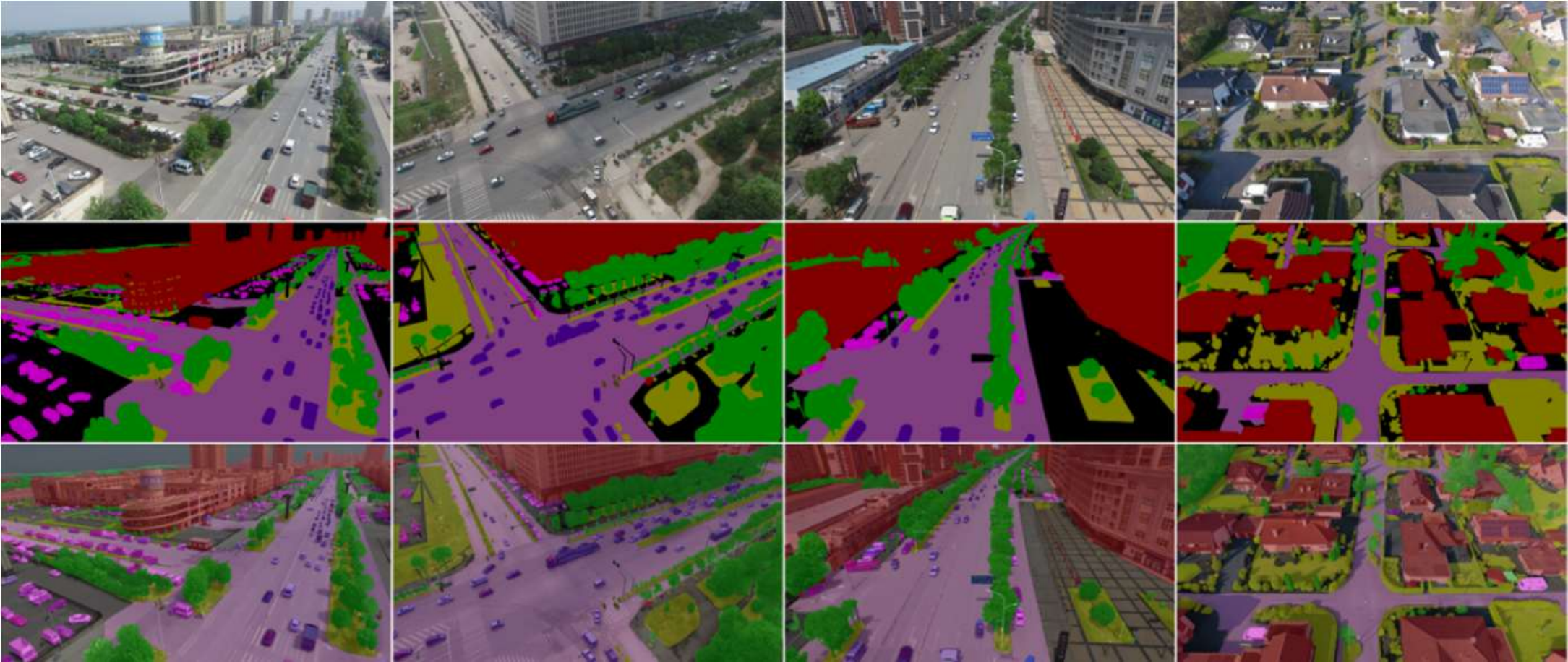}
\caption{For each scene, the top row shows the original aerial RGB image. The middle row provides the corresponding pixel-level ground truth annotation provided by the dataset. The bottom row displays the segmentation masks laid on top of the RGB images.}
\label{fig:UAVid Examples}
\end{figure}

Our first benchmark is the UAVid dataset \citep{lyu2020uavid}, a benchmark for aerial urban scene segmentation. This dataset comprises high-resolution video sequences (up to 4K) captured by UAVs, featuring dense, pixel-level annotations across \textbf{8 distinct classes} such as buildings, roads, vegetation, and moving vehicles. The cluttered urban landscapes of UAVid, illustrated in Figure \ref{fig:UAVid Examples}, introduce significant semantic and visual complexity. The distinction between static and dynamic elements is often little and occurs at a fine spatial scale, providing a rigorous test for the core mechanism of our temporal reuse strategy.

For TTR to succeed here, its lightweight change detection must be sensitive enough to identify small dynamic objects (e.g., individual cars) while robustly classifying large static structures for caching, even under varying lighting. Demonstrating efficiency gains on UAVid validates that TTR's fine-grained, patch-level operation is effective in complex, real-world aerial monitoring scenarios, confirming its applicability for tasks like urban traffic analysis and infrastructure monitoring.

\subsubsection{The A2D2 Dataset}

To push our framework to test beyond the aerial domain, we extend our evaluation to the Audi Autonomous Driving Dataset (A2D2) \citep{geyer2020a2d2}. This large-scale dataset provides video from a forward-facing vehicle camera, annotated with over \textbf{15 semantic classes}. The inclusion of A2D2 represents a shift in our evaluation, moving from a top-down, relatively stable viewpoint to a ground-level perspective characterized by near-constant camera motion. This continuous forward movement induces a global optical flow where nearly every pixel shifts between frames, creating a challenging low-redundancy environment that directly tests the limits of temporal caching.

% FIGURE 5: A2D2 Examples
\begin{figure}[!t]
\centering
\includegraphics[width=\columnwidth]{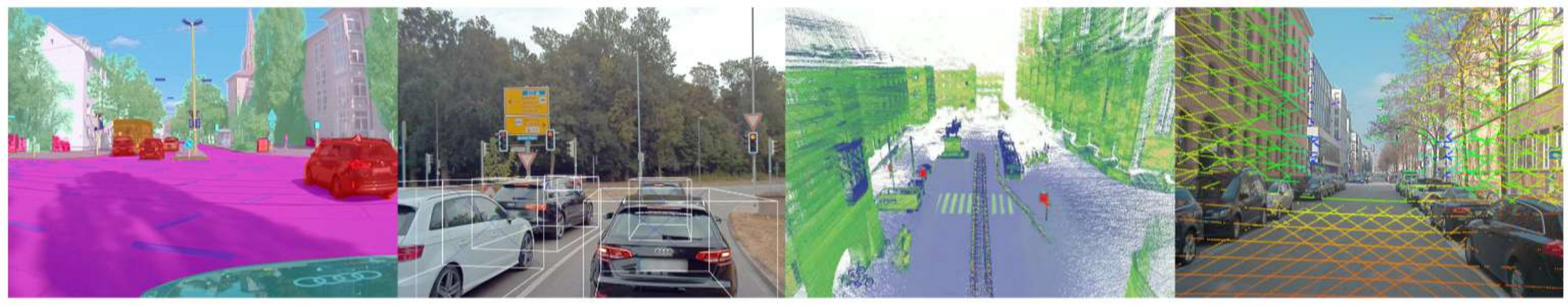}
\caption{Visualizations of A2D2 data. From left: semantic segmentation, 3D bounding boxes, dense point cloud from SLAM, single frame point cloud overlaid on corresponding camera image}
\label{fig:A2D2 Examples}
\end{figure}

This scenario serves as a crucial stress test for the sensitivity of our similarity metric. Demonstrating a significant performance boost on A2D2 (see Figure \ref{fig:A2D2 Examples}) would prove that TTR is not just an optimization for static scenes but a robust, general-purpose technique. It validates that our approach can adaptively exploit whatever degree of temporal redundancy exists in a given video stream, making it a viable solution for a broader class of mobile robotics and vision applications where camera motion is persistent.

\subsubsection{The FloodNet Dataset}
To validate our method against established community standards for post-disaster assessment, we incorporate the \textbf{FloodNet} dataset. FloodNet is a high-resolution aerial imagery benchmark captured after Hurricane Harvey, containing precise annotations for classes such as "Flooded Building," "Flooded Road," and "Non-Flooded Building." While FloodNet is primarily an image-based benchmark, applying our video-based TTR framework to simulated sequences from this dataset allows us to verify that our method performs robustly on the specific semantic classes required for damage assessment, ensuring that the efficiency gains of TTR do not come at the cost of recognizing critical disaster features.

\subsubsection{The Floodwater Dataset}

% FIGURE 6: Floodwater Examples
\begin{figure}[!t]
\centering
\includegraphics[width=\columnwidth]{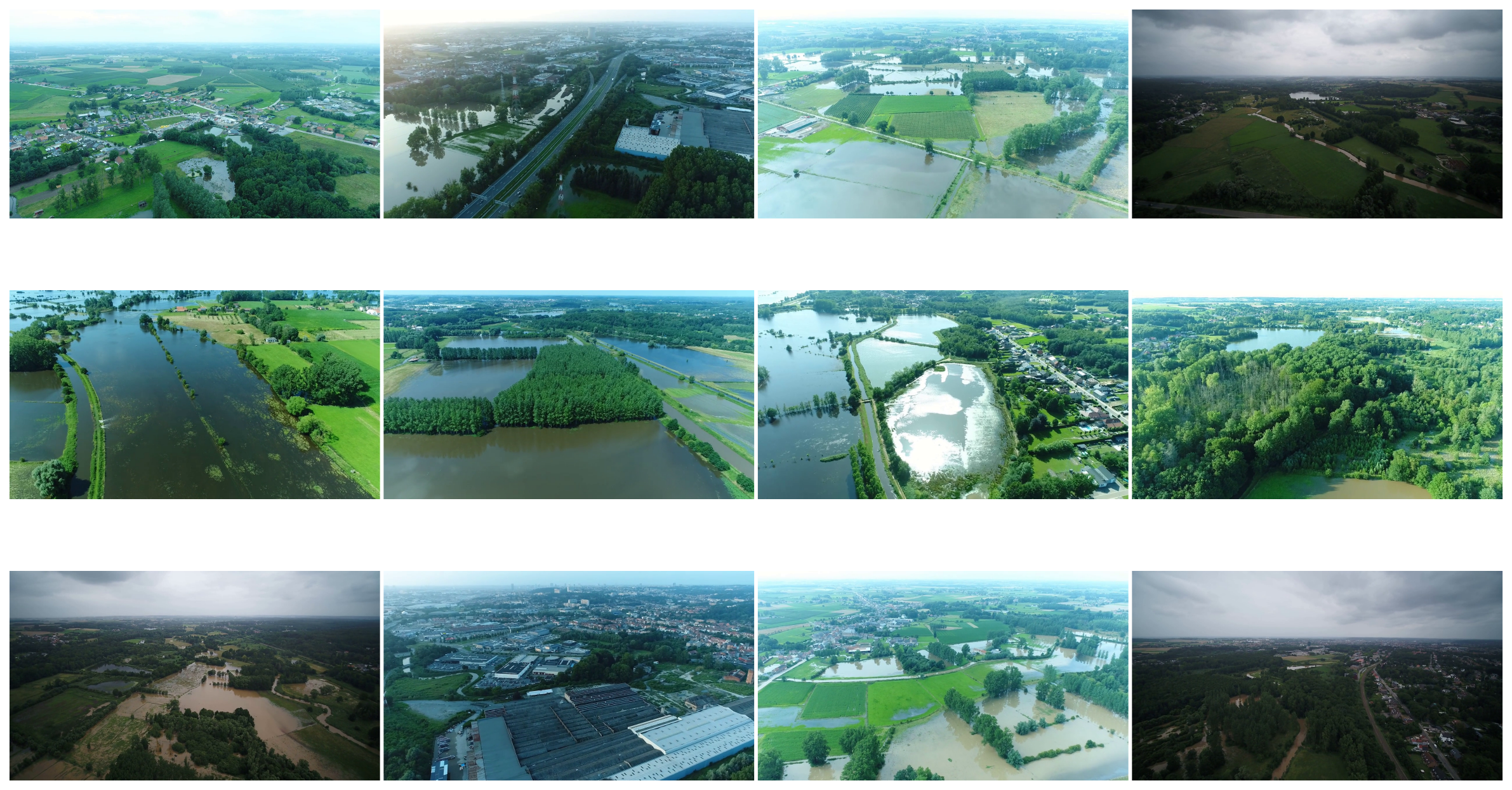}
\caption{Visual diversity and challenges within the Floodwater dataset. This selection of frames highlights the complex scenarios for aerial semantic segmentation, ranging from flooded urban infrastructure (top-left, bottom-middle) to inundated agricultural fields (middle row). The dataset captures varying environmental conditions, from clear, sunny days with reflections on the water to overcast and stormy weather (top-right, bottom-left). This validates the dataset as a benchmark for developing and testing real-world flood monitoring systems.}
\label{fig:Floodwater Examples}
\end{figure}

Floodwater segmentation is an ideal example of a real-life application for which TTR is relevant. The information needs to be available as fast as possible, while resource consumption should be minimized in order to maximize the flight time and thus the area covered in a single flight.
A significant challenge in developing and benchmarking robust flood mapping algorithms for UAVs is the scarcity of large-scale, publicly available video datasets with dense semantic annotations.
While several existing UAV-based datasets are available \citep{DBLP:journals/corr/abs-2012-02951, 10.1007/978-3-031-84602-1_14, 10404232, 10737086}, they are often limited in scope for our specific use case. These limitations include low resolution, a lack of diversity in water conditions (e.g., varying turbidity, reflections), (nearly) perpendicular camera angles, though in practice, drones are mostly flown in oblique viewing mode, and critically, a scarcity of dense, frame-by-frame annotations for continuous video data. The process of manual, frame-by-frame semantic annotation for high-resolution video is notoriously labor-intensive, time-consuming, and cost-prohibitive, which explains the current data gap.

To address this gap and to provide a realistic benchmark for our TTR framework, we have curated a new dataset, which we term the Floodwater Dataset (see Figure \ref{fig:Floodwater Examples}). We collected \textbf{30 distinct video sequences} of flood-affected regions in Flanders, Belgium, from publicly available sources. These videos range from \textbf{2 to 20 minutes} in duration and were captured at high resolutions varying from \textbf{720p to 4K}, providing the fine-grained detail necessary for precise segmentation. The sequences were recorded at different flying speeds, resulting in a mix of slow-panning and fast-changing imagery, which allows for a robust evaluation of our model's ability to adapt to varying levels of temporal change.

To overcome the annotation bottleneck, we leverage the Segment Anything Model 2 (SAM2) \citep{ravi2024sam2segmentimages}, a state-of-the-art foundation model for interactive segmentation, within a semi-supervised framework to generate high-quality pseudo-labels. The process begins with minimal human input: on a few keyframes of a video sequence, a user simply provides a few positive and negative point prompts to identify the primary water bodies that are to be tracked. These pre-annotated frames are then passed to SAM2, which leverages its video segmentation and tracking capabilities to automatically propagate the segmentation masks throughout the rest of the video sequence. This method efficiently generates a reliable and high-quality ground truth of frame-based segmentation masks for training and evaluation, turning our diverse but unlabeled video collection into a valuable, ready-to-use benchmark dataset.

The Floodwater dataset serves as the focus benchmark for our research. Its combination of high-resolution video and significant temporal redundancy (i.e., large static areas in slow-panning shots) creates the exact computational challenge that TTR is designed to solve. It allows us to directly measure performance gains in a realistic application and validate the suitability of our approach for real-time flood detection and damage assessment on resource-constrained UAVs. A key contribution of this dataset is the generation of dense, pixel-level pseudo-labels for the critical binary classification of flooded versus non-flooded areas in video sequences. To foster further research in this critical domain, the dataset is made available at \url{https://github.com/decide-ugent/floodwater-dataset}.

\subsection{Baseline and Experimental Architectures}

To evaluate the performance and versatility of our proposed Temporal Token Reuse (TTR) framework, we establish a set of baselines representing diverse architectural paradigms and select standard backbones for integration. Our evaluation strategy is twofold: first, we compare our final TTR-enhanced models against existing segmentation networks; second, we integrate TTR into common backbones to demonstrate its direct impact and compatibility.

For our baseline comparisons, we select three architectures. First, we include SegFormer \citep{xie2021segformer} as a representative of the current state-of-the-art Transformer-based models. Given the dominance of Transformers in many computer vision tasks, it is crucial to benchmark our CNN-based approach against a leading Transformer to contextualize its performance within the broader landscape of modern segmentation networks. Second, to represent architecturally efficient designs, we select MobileNetV2 \citep{Sandler2018MobileNetV2} and BiSeNet \citep{yu2018bisenet}. These models serve as critical baselines as they are highly optimized for real-time performance on resource-constrained hardware and are widely adopted in edge applications, the primary target domain for our work.

To demonstrate the direct impact and architectural compatibility of our TTR mechanism, we integrate it into two widely adopted and foundational CNN backbones: ResNet \citep{he2016deep} and EfficientNet \citep{tan2019efficientnet}. The choice of these backbones is deliberate. Our patch-based caching and reuse strategy is designed to operate on feature maps produced by standard convolutional layers. Architectures that heavily rely on specialized operators like atrous (dilated) convolutions to expand their receptive field, such as DeepLab\citep{DBLP:journals/corr/ChenPK0Y16}, can disrupt the direct spatial correspondence between image patches and their resulting feature representations, which would complicate our feature sharing between patches. ResNet and EfficientNet, with their conventional convolutional structures, provide a clean and direct testbed to isolate and measure the efficiency gains attributable solely to our temporal reuse strategy.

\subsection{Quantitative Performance}

In the following sections, we present a quantitative analysis of segmentation accuracy, measured by mean Intersection over Union (mIoU) and Pixel Accuracy, and computational efficiency, measured in Frames Per Second (FPS). All models were benchmarked on two hardware platforms: a high-end NVIDIA GeForce GTX 1080 Ti for desktop performance and a power-constrained NVIDIA Jetson Orin Nano to simulate a real-world edge deployment scenario. The models enhanced with our method are denoted with ``(TTR)''. This allows for a clear evaluation of TTR's effectiveness on standard, off-the-shelf architectures while comparing the final, optimized models against leading purpose-built and Transformer-based solutions.

\subsubsection{Performance on Floodwater \& FloodNet Datasets}

The performance of our TTR framework on the hydrological datasets, detailed in Table \ref{tab:flood_performance}, confirms both its operational efficiency and semantic robustness.

On the \textbf{Floodwater Dataset}, which represents our primary target application for real-time monitoring, TTR delivers substantial speed gains. This dataset is characterized by large, static regions of land and water, providing an ideal scenario for exploiting temporal redundancy. When applied to EfficientNet-B4, as seen in Table~\ref{tab:flood_performance} TTR increases inference speed by a factor of 1.67x on both the high-end GTX 1080 Ti (from 30 to 50 FPS) and the resource-constrained Jetson Orin Nano (from 15 to 25 FPS). This significant acceleration is achieved with a minimal fidelity trade-off, incurring a negligible mIoU drop of only 0.4 percentage points. The impact on the Jetson platform is particularly critical; by elevating a high-accuracy model like EfficientNet-B4 from a non-real-time 15 FPS to a deployable 25 FPS, TTR pushes it into a viable real-time operational window for edge devices.

Complementing this, the results on the \textbf{FloodNet} benchmark validate the method's robustness against community standards for damage assessment. While FloodNet is a challenging dataset with fine-grained class distinctions (e.g., distinguishing "Flooded Road" from "Non-Flooded Road"), TTR demonstrates remarkable stability. As shown in Table \ref{tab:flood_performance}, the EfficientNet-B4 + TTR model maintains the exact same accuracy as the dense baseline (62.1\% mIoU). This result is pivotal: it confirms that the sparsity introduced by our token reuse mechanism does not compromise the recognition of critical disaster features. Even when the efficiency gains are modest as FloodNet sequences often lack the temporal continuity of raw drone flight footage the framework ensures that segmentation quality remains preserved, validating TTR as a safe, reliable choice for mission-critical analysis.

\subsubsection{Performance on UAVid and A2D2 Datasets}

To validate the robustness of our method, we evaluated on established benchmarks that have complex, real-world conditions. We selected the UAVid dataset, a prominent academic benchmark for aerial urban scene understanding. Featuring 8 distinct classes in diverse environments, this dataset challenges the framework to correctly distinguish between static infrastructure and dynamic objects like moving vehicles from an aerial perspective.

The results in Table~\ref{tab:other_performance} confirm that the benefits of TTR generalize effectively to this multi-class, real-world-inspired environment. On the 1080 Ti, EfficientNet-B4 with TTR jumps from 32 FPS to 54 FPS (a 1.69x speedup), while the mIoU decreases by only 0.3\%. A consistent speedup is also observed on the Jetson Orin Nano, where the frame rate is boosted from 16 to 27 FPS. This demonstrates the validity of our lightweight similarity metric, which effectively identifies static elements like buildings and roads for caching, thereby confirming TTR's applicability to a wide range of aerial segmentation tasks.

Next, to stress-test our approach in a fundamentally different and more demanding context, we transitioned from an academic aerial benchmark to a dataset representing a direct, complex real-world application: autonomous driving. We utilized the A2D2 dataset, which captures the challenges of constant ego-motion and dense traffic from a ground-level view. This scenario presents a significant challenge, as the proportion of the scene that is truly static is much smaller than in typical aerial footage.

% TABLE 1: Floodwater and FloodNet
\begin{table*}[!t]
\centering
\caption{Performance on Flood-related Datasets (Floodwater and FloodNet). Accuracy metrics are consistent across hardware, while FPS is reported for both GTX 1080 Ti and Jetson Orin Nano.}
\label{tab:flood_performance}
\setlength{\tabcolsep}{6pt}
\begin{tabular*}{\textwidth}{@{\extracolsep{\fill}} llcccc}
\toprule
& & & & \multicolumn{2}{c}{\textbf{Inference Speed}} \\
\cmidrule(l){5-6}
\textbf{Dataset} & \textbf{Model} & \textbf{mIoU (\%)} & \textbf{Pix. Acc. (\%)} & \textbf{FPS (1080 Ti)} & \textbf{FPS (Orin Nano)} \\
\midrule
\multirow{7}{*}{\rotatebox{90}{\textbf{Floodwater}}}
& SegFormer-B0             & 82.0 & 92.0 & 25 & 12 \\
& EfficientNet-B4          & 78.0 & 88.0 & 30 & 15 \\
& \textbf{EfficientNet-B4 + TTR} & 77.6 & 87.8 & 50 & 25 \\
& ResNet-50                & 74.0 & 83.0 & 22 & 11 \\
& \textbf{ResNet-50 + TTR} & 73.8 & 82.6 & 35 & 18 \\
& BiSeNetV2                & 70.0 & 81.0 & 65 & 32 \\
& MobileNetV3              & 68.0 & 76.0 & 80 & 40 \\
\midrule
\multirow{7}{*}{\rotatebox{90}{\textbf{FloodNet}}}
& SegFormer-B0             & 66.5 & 87.2 & 25 & 12 \\
& EfficientNet-B4          & 62.1 & 84.5 & 30 & 15 \\
& \textbf{EfficientNet-B4 + TTR} & 62.1 & 84.5 & 29 & 14 \\
& ResNet-50                & 56.5 & 80.4 & 22 & 11 \\
& \textbf{ResNet-50 + TTR} & 56.5 & 80.4 & 21 & 10 \\
& BiSeNetV2                & 54.2 & 79.0 & 65 & 32 \\
& MobileNetV3              & 51.0 & 76.5 & 80 & 40 \\
\bottomrule
\end{tabular*}
\end{table*}

% TABLE 2: UAVid and A2D2
\begin{table*}[!t]
\centering
\caption{Performance on UAVid and A2D2 Datasets. TTR provides significant speedups on these video datasets compared to the baseline models.}
\label{tab:other_performance}
\setlength{\tabcolsep}{6pt}
\begin{tabular*}{\textwidth}{@{\extracolsep{\fill}} llcccc}
\toprule
& & & & \multicolumn{2}{c}{\textbf{Inference Speed}} \\
\cmidrule(l){5-6}
\textbf{Dataset} & \textbf{Model} & \textbf{mIoU (\%)} & \textbf{Pix. Acc. (\%)} & \textbf{FPS (1080 Ti)} & \textbf{FPS (Orin Nano)} \\
\midrule
\multirow{7}{*}{\rotatebox{90}{\textbf{UAVid}}}
& SegFormer-B0             & 62.3 & 86.0 & 25 & 12 \\
& EfficientNet-B4          & 56.0 & 84.6 & 32 & 16 \\
& \textbf{EfficientNet-B4 + TTR} & 55.7 & 84.1 & 54 & 27 \\
& ResNet-50                & 53.8 & 80.1 & 25 & 12 \\
& \textbf{ResNet-50 + TTR} & 53.2 & 79.8 & 40 & 20 \\
& BiSeNetV2                & 52.0 & 81.0 & 65 & 32 \\
& MobileNetV3              & 50.2 & 77.2 & 71 & 35 \\
\midrule
\multirow{7}{*}{\rotatebox{90}{\textbf{A2D2}}}
& SegFormer-B0             & 65.0 & 85.0 & 18 & 9 \\
& EfficientNet-B4          & 60.0 & 80.0 & 25 & 13 \\
& \textbf{EfficientNet-B4 + TTR} & 59.7 & 79.4 & 42 & 21 \\
& BiSeNetV2                & 58.0 & 80.0 & 55 & 28 \\
& ResNet-50                & 54.0 & 74.0 & 24 & 12 \\
& \textbf{ResNet-50 + TTR} & 53.8 & 73.6 & 38 & 19 \\
& MobileNetV3              & 52.0 & 81.0 & 70 & 35 \\
\bottomrule
\end{tabular*}
\end{table*}

Even in this highly dynamic environment, TTR delivered substantial efficiency gains. The EfficientNet-B4 model was accelerated from 25 to 42 FPS on the 1080 Ti (1.68x speedup) and from 13 to 21 FPS on the Orin Nano (1.62x speedup), all with a minimal mIoU drop of only 0.3\%. The fact that a significant speedup was still achieved proves that TTR is not confined to scenarios with vast static backgrounds. Instead, it robustly adapts to the level of dynamism in the scene, caching features for momentarily static elements (e.g., distant buildings, parked cars) while prioritizing computation for moving ones. This adaptability is critical for real-world deployment.

% FIGURE 7: Accuracy vs Efficiency
\begin{figure}[!t]
\centering
\subfloat[]{\includegraphics[width=2.5in]{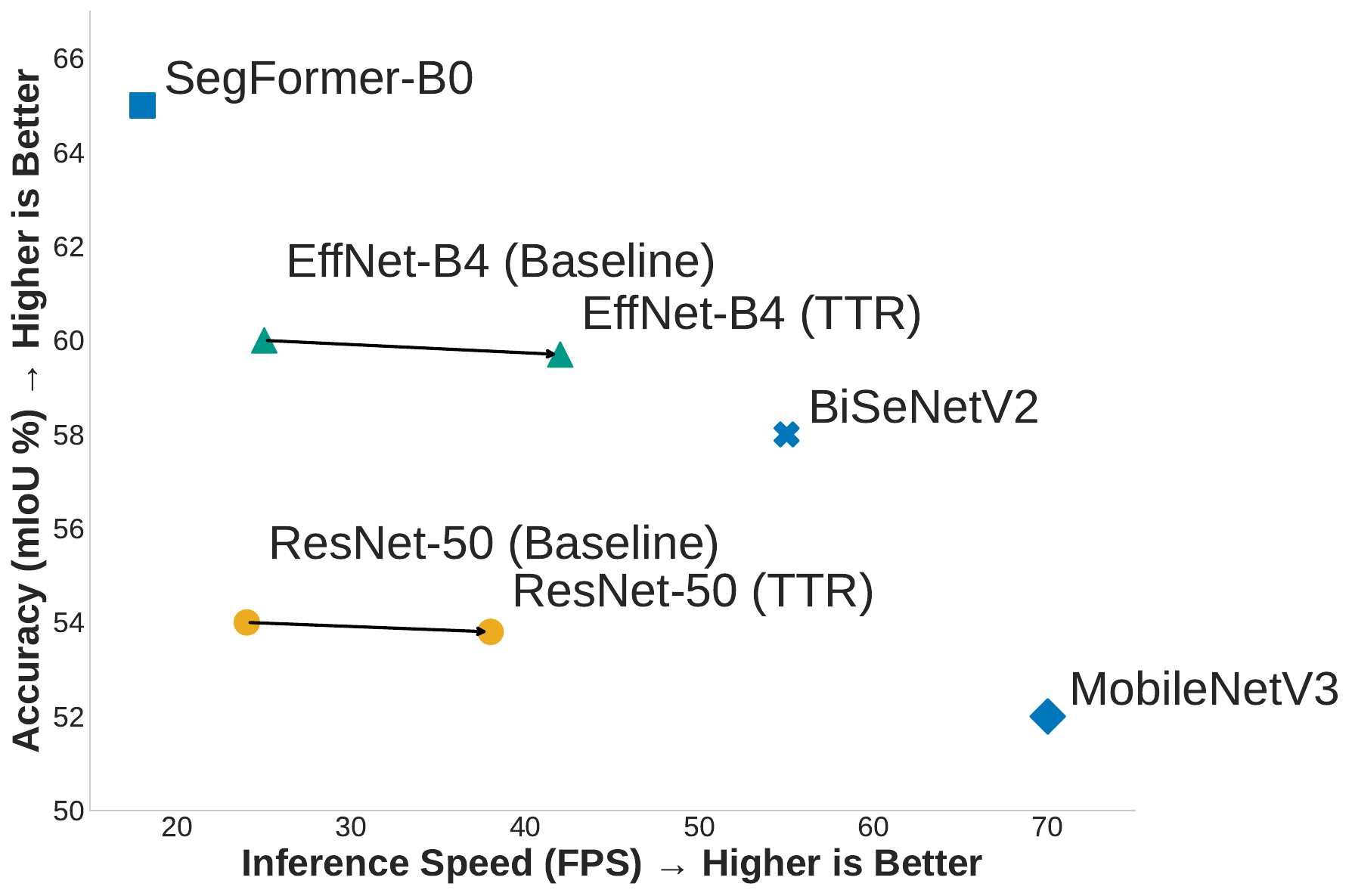}%
\label{fig:accuracy_vs_efficiency}}
\hfil
\subfloat[]{\includegraphics[width=2.5in]{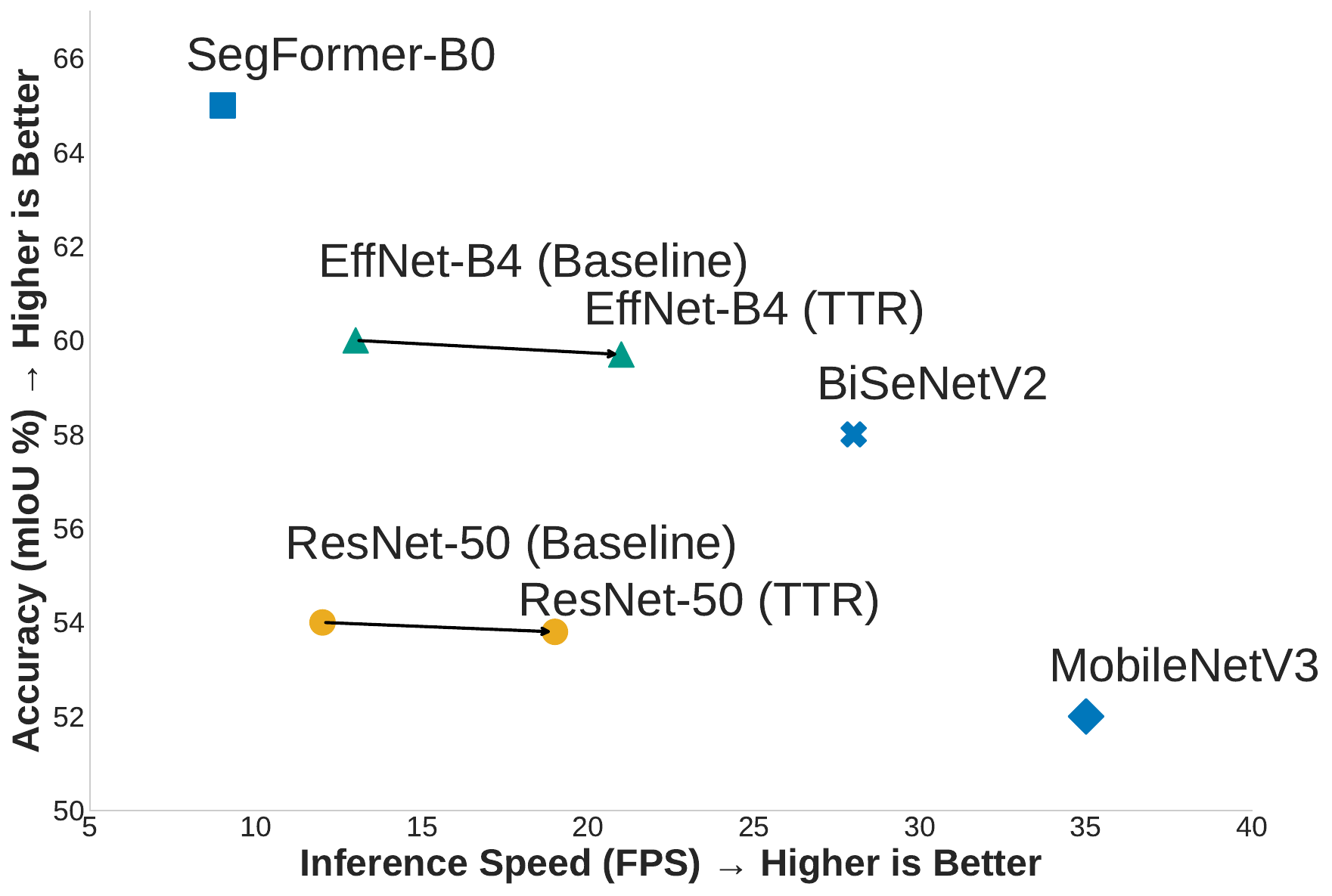}%
\label{fig:sub2}}
\caption{The accuracy-efficiency trade-off on the Floodwater dataset, evaluated on (a) the NVIDIA 1080 Ti and (b) the Jetson Orin Nano. The arrows illustrate the performance shift from baseline models to their TTR-enhanced counterparts, showing a move towards the ideal top-right corner (high accuracy, high speed) and confirming a significant efficiency gain for a negligible accuracy cost.}
\label{fig:accuracy_vs_efficiency_2}
\end{figure}

The overall impact of TTR on the accuracy-efficiency trade-off across these datasets is visualized for the Nvidia 1080 Ti and a Jetson Orin Nano in Figure~\ref{fig:accuracy_vs_efficiency} and Figure~\ref{fig:sub2}, respectively. The plots clearly illustrate how TTR shifts the performance of both ResNet-50 and EfficientNet-B4 towards the ideal top-right corner. Our method enables these more accurate, heavier backbones to achieve FPS rates competitive with much lighter models, creating a new and more favorable Pareto frontier for real-time segmentation on edge devices.

\section{Discussion}\label{sec:discussion}

The results presented consistently validate our central hypothesis: exploiting temporal redundancy via token reuse offers an improvement in computational efficiency for a minimal and often negligible cost in accuracy. Across all datasets and hardware platforms, we observed an increase in processing speed, a crucial factor for real-time deployment on resource-constrained platforms like UAVs.

The success of TTR is rooted in the inherent spatio-temporal redundancy of video data, particularly in aerial footage. The strong performance on the Floodwater and UAVid datasets underscores how well our method is suited to these scenarios, where large portions of the frame are often static or slow-moving. By intelligently identifying and skipping redundant computations for these regions, TTR can allocate the UAV's limited processing power where it is most needed. The choice of cosine similarity as the change detection metric further enhances this robustness by making the system less susceptible to global illumination shifts a frequent challenge in outdoor environments that can confound simpler difference-based methods.

% FIGURE 8: Optical Flow / Computational Load
\begin{figure}[!t]
\centering
\includegraphics[width=\columnwidth]{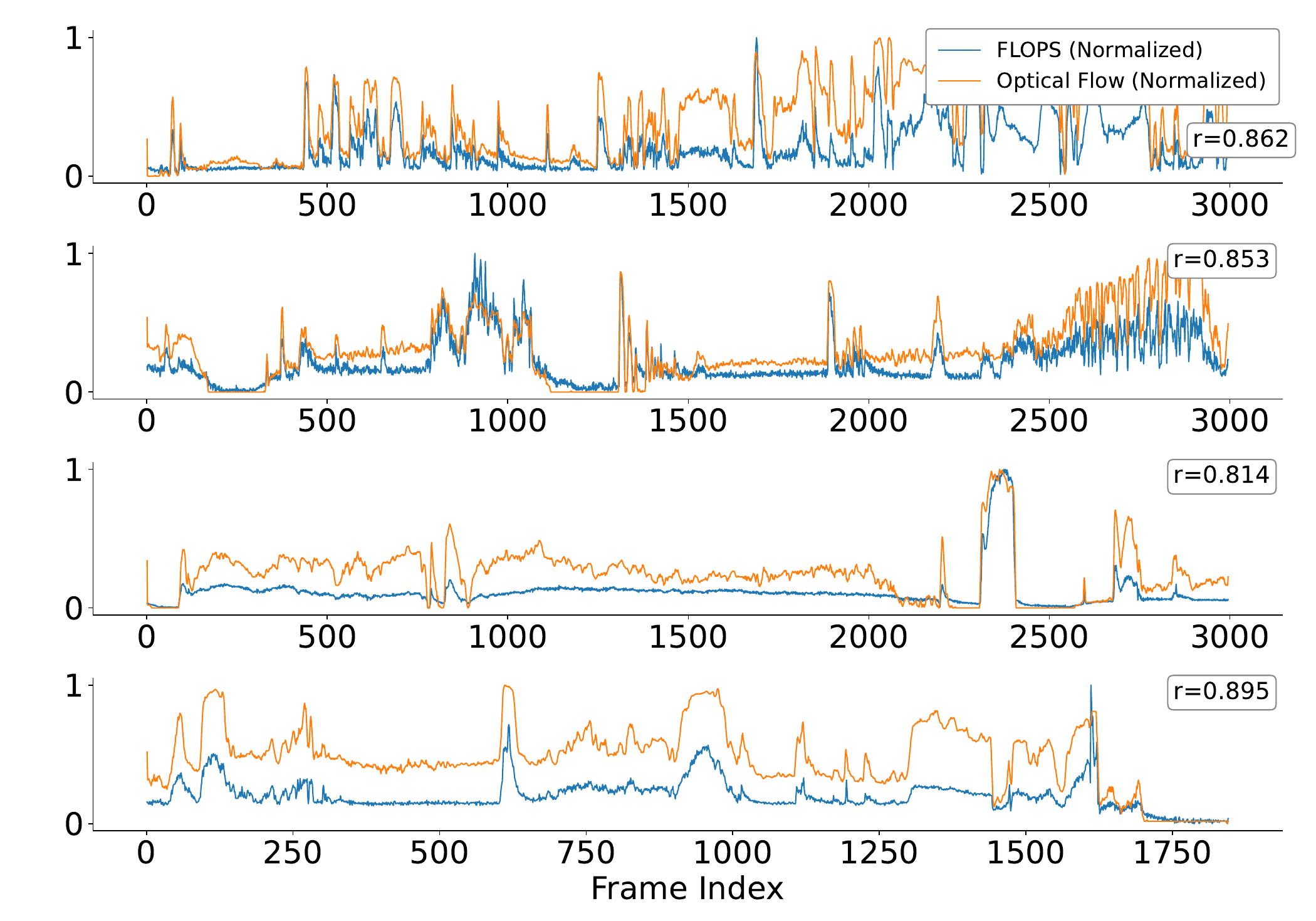}
\caption{Correlation between scene dynamism and computational load of our TTR framework. The plots compare the normalized computational load (FLOPS, blue) of our TTR-enhanced model against the normalized magnitude of inter-frame optical flow (orange) for four representative video sequences from the Floodwater dataset. The strong visual alignment and high Pearson correlation coefficients (r > 0.8) demonstrate that TTR effectively adapts its processing, allocating more computational resources during periods of high motion and significantly reducing computation when the scene is static.}
\label{fig:of_impact}
\end{figure}

To quantitatively validate this adaptive behavior, we analyze the direct relationship between scene dynamism and the computational load of our TTR-enhanced model in Figure \ref{fig:of_impact}. In this analysis, we use the magnitude of inter-frame optical flow as a proxy for scene dynamics; low values signify a static scene (e.g., a hovering UAV), while high values indicate significant motion from either the camera's ego-motion or objects within the scene. The plots reveal a remarkably tight coupling between the normalized FLOPs executed by our model and the optical flow. The high Pearson correlation coefficients (r > 0.8) demonstrate a causal link, that our lightweight similarity metric is an effective and efficient proxy for complex motion estimation, successfully guiding the model's computational budget in real time.

The patterns depicted in the graph provide insights into the video behavior. The troughs, where the computational load drops significantly, correspond to periods of minimal change, such as when the UAV is hovering or panning slowly over a large, uniform body of water. In these moments, TTR achieves maximum efficiency by reusing the vast majority of feature patches from the previous frame. Conversely, the sharp peaks in both optical flow and FLOPs signify moments of high dynamism. These can be triggered by sudden UAV movements like sharp turns or rapid altitude changes, or by the appearance of dynamic objects. The fact that TTR immediately scales up its computation in response to these events is critical, as it confirms the framework’s ability to preserve high-fidelity processing when accuracy is important. This dynamic scaling ensures that TTR is not just an efficiency-saving mechanism but a truly adaptive one that aligns its effort with the temporal complexity of the scene.

% FIGURE 9: Qualitative Results
\begin{figure}[!t]
\centering
\includegraphics[width=\columnwidth]{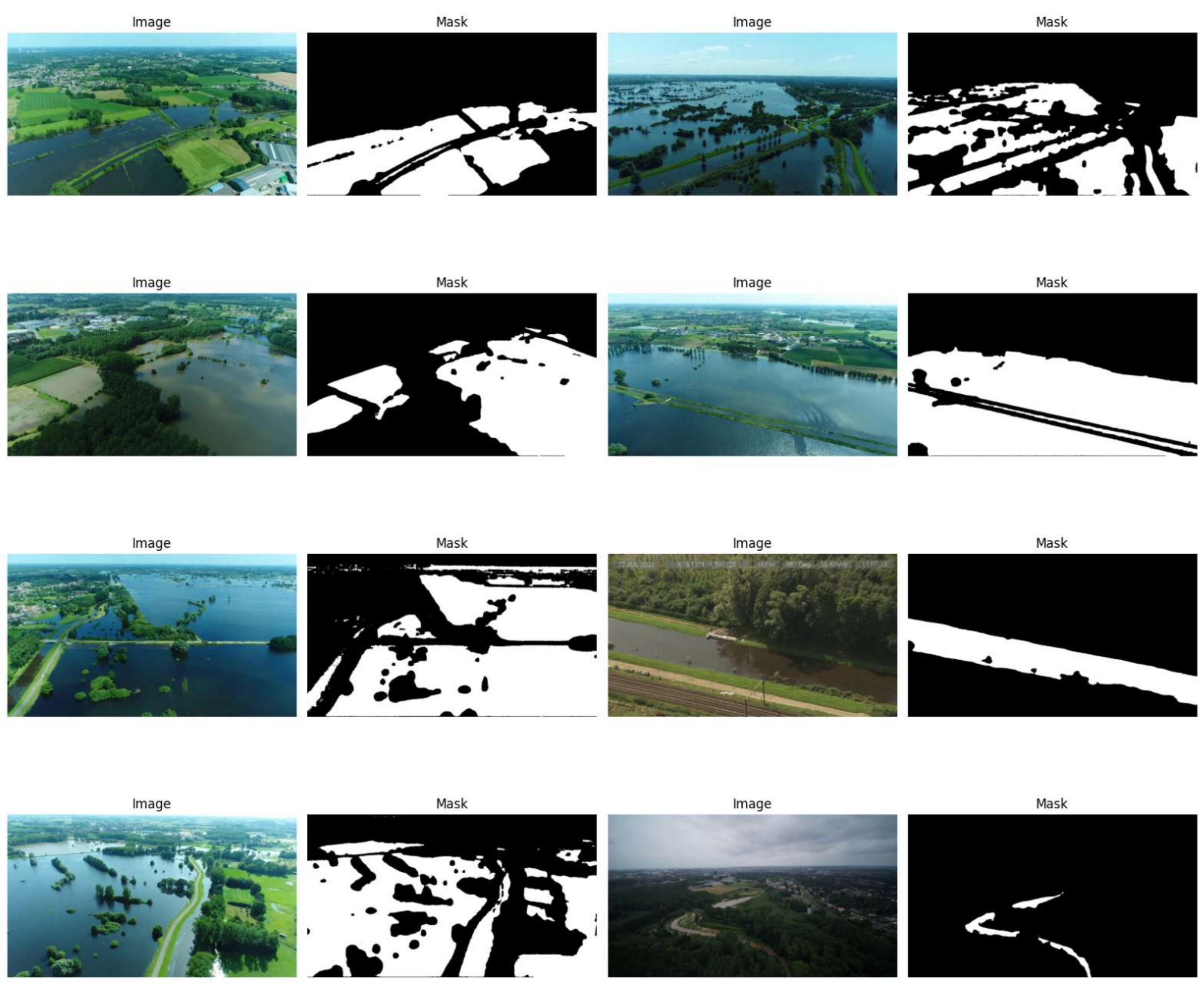}
\caption{Visualization of segmentation performance on representative floodwater scenes: left input images; right predicted masks}
\label{fig:result_labels}
\end{figure}

Crucially, the metrics in Figure \ref{fig:of_impact} emerge from TTR's fine-grained, patch-level decision-making. TTR excels in these low-motion scenarios by independently evaluating each patch, a capability that allows it to focus resources on dynamic regions while saving computation on static ones. This patch-wise adaptability is an advantage over coarse, frame-level methods that process the entire frame if any part of it changes. Importantly, these efficiency gains do not compromise performance. The qualitative analysis in Figure \ref{fig:result_labels} provides visual confirmation, showing that the predicted masks remain accurate even in complex scenes.

This adaptive resource allocation is important for real-world deployment on UAVs, where energy and onboard compute are strictly limited. By changing its computational footprint in direct proportion to scene complexity, TTR achieves substantial energy savings during common, low-motion periods of operation, extending mission duration. This behavior makes it a robust and reliable solution, ensuring that the system can handle bursts of activity without failure while remaining maximally efficient during low-motion phases, especially for UAVs or hardware-constrained applications.

\section{Conclusion}\label{sec:conclusion}

In this paper, we introduced Temporal Token Reuse (TTR), a framework designed to significantly improve the computational efficiency of video semantic segmentation, with a particular focus on UAV applications. By treating image patches as tokens and reusing the features of those tokens that remain static over time, TTR directly addresses the problem of temporal redundancy inherent in video streams. Our approach successfully marries the spatial adaptability of patch-based CNNs with a new layer of temporal intelligence.

Our primary contribution is the development of the Temporal Token Reuse (TTR) framework and the demonstration of its effectiveness in accelerating video semantic segmentation without substantially compromising accuracy. TTR introduces a principled mechanism for identifying and reusing temporally redundant token features through a lightweight cosine-similarity metric and a deep feature cache. By integrating this temporal reuse strategy directly into a patch-based CNN pipeline, the framework enables substantial gains in inference speed (up to 30\% faster) while incurring only a marginal loss in segmentation accuracy. This work validates that it is not always necessary to re-process all spatial content in every frame to maintain high-quality semantic understanding.

Ultimately, TTR represents a significant step towards enabling powerful, real-time perception on autonomous systems, making sophisticated computer vision models more accessible for a new generation of intelligent edge devices.

\section{Limitations and Future Research}\label{sec:limitations}

While Temporal Token Reuse (TTR) significantly reduces the computational latency of on-board segmentation, bridging the gap between pixel-level inference and actionable geospatial intelligence presents several opportunities for future exploration.

A primary limitation of the current framework is that it operates exclusively in the camera's image plane. While oblique imagery provides superior visibility of vertical infrastructure and water levels compared to nadir views, the resulting segmentation masks are inherently subject to perspective distortion and varying scale. For crisis responders, a segmented pixel mask provides qualitative awareness but lacks the direct geospatial context required to assess the precise impact on specific infrastructure.

A promising avenue for future research involves the integration of real-time \textbf{orthorectification} directly into the edge inference pipeline. By fusing the UAV's telemetry data (IMU and GPS) with a Digital Elevation Model (DEM), future systems could project these oblique segmentation masks onto a georeferenced cartographic plane during flight. This projection would be a critical prerequisite for overlaying flood extents onto standard topographic maps, enabling the automated analysis of \textbf{real-time road network connectivity}. Such a workflow would allow the system to autonomously identify severed transit routes and calculate accessible paths for emergency vehicles, thereby directly linking computer vision outputs to the operational needs of crisis response.

Additionally, further performance gains could be realized by exploring adaptive mechanisms for the similarity threshold $\tau$. While a fixed threshold proved robust in our experiments, developing a learnable gating policy that adjusts to environmental factors (e.g., flight altitude or weather conditions) could dynamically optimize the trade-off between sparsity and accuracy without manual calibration.

\section*{Acknowledgments}
The authors would like to express their gratitude to the Flanders Environment Agency (VMM) and VITO (Flemish Institute for Technological Research) for providing the extensive aerial video dataset that was instrumental to this study. This research was funded by the Belgian Science Policy Office (BELSPO) via the Stereo IV Program, grant number SR/00/415 (FLOWS). S. Leroux and P. Simoens acknowledge the support of the Flanders AI Research Program.

\section*{Declarations}
\begin{itemize}
\item \textbf{Conflict of interest/Competing interests:} The authors declare no competing interests.
\item \textbf{Ethics approval and consent to participate:} Not applicable.
\item \textbf{Consent for publication:} Not applicable.
\item \textbf{Data availability:} The A2D2, UAVid and FloodNet datasets are publicly available. The Floodwater dataset is shared.
\item \textbf{Materials availability:} Not applicable.
\item \textbf{Code availability:} The code is available at \url{https://github.com/decide-ugent/adaptive-segmentation}.
\end{itemize}

\bibliographystyle{IEEEtran}
\bibliography{bibliography}

\begin{IEEEbiography}[{\includegraphics[width=1in,height=1.25in, clip, keepaspectratio]{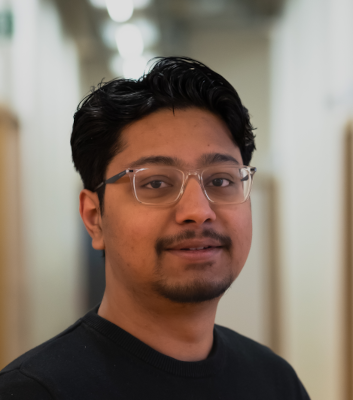}}]{Vishisht Sharma}
 is currently pursuing the Ph.D. degree with the IDLab, Department of Information Technology, Ghent University - imec, Belgium, and VITO (Flemish Institute for Technological Research), Mol, Belgium. 
 
 His research focuses on efficient deep learning and Adaptive Computation.
\end{IEEEbiography}

\begin{IEEEbiography}[{\includegraphics[width=1in,height=1.25in,clip, keepaspectratio]{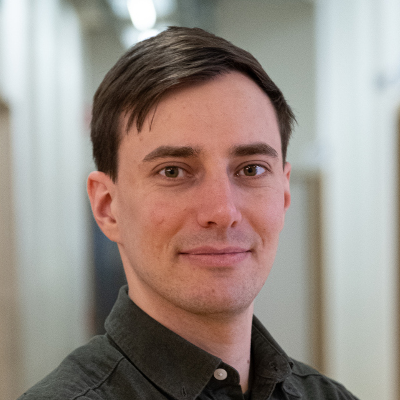}}]{Sam Leroux}
is an assistant professor at the IDLab, an IMEC research group at Ghent University, Ghent, Belgium. 

His main research interests include efficient neural network architectures and edge computing.
\end{IEEEbiography}

\begin{IEEEbiography}[{\includegraphics[width=1in,height=1.25in,clip, keepaspectratio]{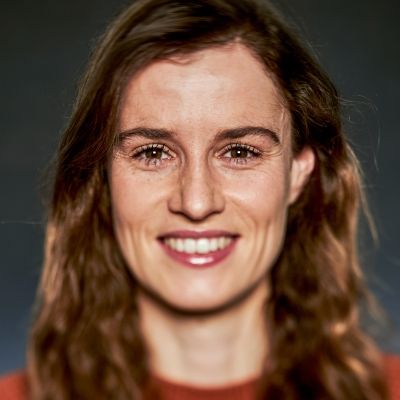}}]{Lisa Landuyt}
 has been a Remote Sensing Scientist with VITO Remote Sensing, Mol, Belgium. 
 
Her research leverages Machine Learning and remote sensing to enhance the interpretation of satellite imagery for improved environmental monitoring and decision-support.
\end{IEEEbiography}

\begin{IEEEbiography}[{\includegraphics[width=1in,height=1.25in,clip, keepaspectratio]{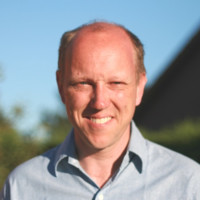}}]{Nick Witvrouwen}
is a Researcher at VITO (Flemish Institute for Technological Research), Mol, Belgium. 

His research interests lie at the intersection of hardware and software for remote sensing, with a specific focus on onboard data processing, edge AI implementation on embedded platforms and the development of efficient data compression algorithms.
\end{IEEEbiography}

\begin{IEEEbiography}[{\includegraphics[width=1in,height=1.25in,clip, keepaspectratio]{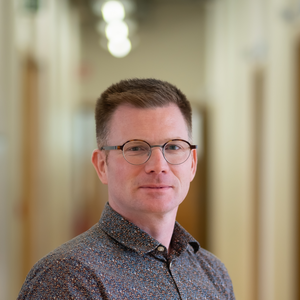}}]{Pieter Simoens}
currently holds a position as Associate Professor at Ghent University, Ghent, Belgium, and is also affiliated with imec. 

His main research interests include the domain of intelligent distributed systems, with a specific focus on resource-efficiency, unsupervised learning, and collective intelligence in large-scale networks.
\end{IEEEbiography}

\vfill

\end{document}